\pdfoutput=1

\documentclass[11pt]{article}

\usepackage{EMNLP2023}

\usepackage{times}
\usepackage{latexsym}

\usepackage{adjustbox}

\usepackage[T1]{fontenc}

\usepackage[utf8]{inputenc}

\usepackage{microtype}

\usepackage{inconsolata}

\title{Measuring a hate speech spectrum with faceted Rasch item response theory and perspective-aware, explainable-by-design deep learning}

\usepackage{graphicx}

\author{  Chris J. Kennedy \\ 
  Center for Precision Psychiatry, Mass General Hospital \\
  Department of Psychiatry, Harvard Medical School\\
  Boston, MA \\
  \AND
  Geoff Bacon \and Alexander Sahn \and Claudia von Vacano
  D-Lab\\
  University of California, Berkeley\\
  Berkeley, CA \\
}
\begin{document}
\maketitle
\begin{abstract}
We propose a system for measuring hate speech on a continuous, interval-valued spectrum ranging from genocidal to supportive speech by combining supervised deep learning with faceted Rasch item response theory (IRT). We decompose the theoretical construct of hate speech into constituent concepts operationalized as 10 ordinal labels. Those labels are reconstituted via IRT probabilistic latent modeling into an interval outcome measure while simultaneously estimating and adjusting for each annotator's labeling perspective. Our scaling procedure integrates naturally with a multitask deep learning architecture for automated prediction, allowing design-based explainability of the continuous score through those components. We apply this method to a new, open source dataset of 50,070 social media comments sourced from YouTube, Twitter, and Reddit, annotated and labeled by 11,143 United States-based Amazon Mechanical Turk workers. Our RoBERTa-based model shows improved accuracy compared to alternative approaches. This system offers a new paradigm for supervised NLP that encourages continuous rather than binary constructs, and design-based incorporation of annotator perspective and model explainability.
\end{abstract}

\section{Introduction}

Developing hate speech annotation instruments, and the ensuing annotation process, is particularly challenging due to the complexity of the \textit{construct}, i.e. the theoretical framework for a characteristic established to provide a consistent understanding across individuals and over time \citep{wilson2004constructing}. In evaluating the consistency of this labeling process across annotators, agreement metrics such as Krippendorff's alpha coefficient are typically found to be lower for hate speech than for simpler constructs such as sentiment \citep{kiela2020hateful}. Hate speech models also struggle with false positives, such as flagging counterspeech as hateful. 

Recent work has sought to unpack the coarseness of a binary label for hate speech, such as through multi-label prediction \citep{bianchi-etal-2022-just} and the suggestion that scales should be employed \citep{fortuna2022directions}.
Rather than \textit{detecting} or \textit{classifying} hate speech as a dichotomous yes/no label, \textit{measuring} hate speech would place a given text observation on an interval spectrum (e.g. a ruler) in which numeric differences are meaningful, similar to physical measurement systems such as temperature \cite{chang2004inventing}. Such a continuous hate speech score would provide an estimate of \textit{magnitude}: genocidal hate speech could be at one extreme, with dehumanizing, hostile, or biased speech being less extreme; neutral or ambiguous language is in middle region of the scale, then counterspeech and supportive speech on the opposite end (i.e. the scale could be \textit{bipolar}).


\noindent \textbf{Contributions}\hspace{1em} We describe a system to construct a continuous (interval-valued) measure of hate speech by decomposing the construct of hate speech into multiple constituent labels. Those labels are reconstituted by applying faceted Rasch scaling, \citep{linacre1989facet} a form of item response theory designed for annotated data, and then used to train a multitask supervised deep learning model that incorporates annotator perspective. The multitask architecture yields \textit{explainability by design}: the predicted hate score can be fully characterized by the model's predictions on the component labels. We applied our method to a newly created dataset of 50,070 social media comments sourced from YouTube, Twitter, and Reddit. Those comments were reviewed by over 11,000 Amazon Mechanical Turk annotators using a network-oriented annotation design that allowed the annotators' perspectives to be statistically estimated. We compared the accuracy of our approach to alternatives.  Our hope is that this work can help spur wider usage of Rasch measurement theory to go beyond simplified discrete variables and instead develop continuous, interval variables across fields that incorporate an understanding of annotator perspective.

\section{Theorization} 

Hate speech has many definitions across academic disciplines, legal and regulatory language, and in common vernacular. When attempting to systematically measure a social concept, the concept interpretation will vary greatly depending on who is doing the labeling: life experience, familiarity with the vernacular of a given discourse community, historical and geographic context all vary across individuals, whether experts or laypeople.  Difficult-to-measure phenomena like hate speech are an ideal application of our method. This requires careful theorization and translation into a construct and then to a measurement instrument. In our work, we propose a continuous scale for hate speech that contains positive identity speech, counterspeech, neutral speech, hate speech, and extreme forms of hate speech; we will explain the values along that continuum. We draw from three main sources in defining hate speech: academic literature on hate speech and American legal doctrine on hate crimes, and content moderation policies of large internet firms who host online conversations. Although we acknowledge that hate crimes are distinct from hate speech, we believe that they are theoretically similar and interconnected. Targeting of a group or an individual on the basis of their membership in a group is common to most definitions of hate speech \citep{sellars2016defining}. A more severe form is speech that establishes the inferiority of a group to others can precede denying that group's humanity, a key condition for justifying violence \citep{matsuda2018words}.  Following \cite{stanton2013ten}, we place speech supporting the systematic killing of a specific group (geocide) as the most severe form of hate speech. While any type of hate speech can foment mass violence, either directly or through more severe forms of hate speech, calls for genocide or extermination are most directly connected to real-world behavior. Table \ref{tab-theory-levels}

\begin{table*}[th!]
\caption{Theorized qualitative levels of the hate speech --- supportive speech spectrum.}

\begin{tabular}{|p{1.2\columnwidth}|p{0.9\columnwidth}|}

\hline
{\bf Level \& description} & {\bf Example comment} \\ \hline
\textbf{5. Genocide}: support for or intention of systematically killing all or a large number of a protected identity group & Every g**k in \#LosAngeles should be deported or killed. \\ \hline
\textbf{4. Violence}: threat or support of physical force or emotional abuse intended to hurt or kill members of a protected identity group & \#whitepower...I'll hang you n****r \\ \hline
\textbf{3. Dehumanization}: depriving a protected group of human-like qualities, such as comparison to an animal, insect, or disease & You're all a bunch of human cockroaches is what you are... You ch**ks... \\ \hline
\textbf{2. Hostility}: unfriendliness or opposition to a protected identity group, such as through slurs, profanity, or insults & Learn the fucking language you fucking useless immigrant. \\ \hline
\textbf{1. Bias}: inclination or preference against a protected identity group, including prejudice & @[NAME] They are arabs. Do you need any other explanations?	\\ \hline
\textbf{0. Neutral}: descriptive or other non-harmful references to identity groups & Go get a job at Dick's Sporting Goods and try to work at being a better person \\ \hline
\textbf{-1. Counterspeech}: response to hate speech that seeks to undermine its impact and standing & No, the chances of a muslim shooting you in America is almost nil. There are over 50K gun deaths every year christian USA... \\ \hline
\textbf{-2. Supportive}: respectful, prideful, or other solidarity-based messaging about a protected identity group(s) & I'm bi. And a good listener if you need a friend \\ \hline 
\end{tabular}%
\label{tab-theory-levels}
\end{table*}

\section{Data}

\subsection{Collection}

Social media comments were collected from three popular platforms: YouTube, Reddit, and Twitter. We chose these platforms for their popularity, as respectively, they are used by 73\%, 22\%, and 11\% of U.S. adults \citep{pew2019socialmedia}. Prior work on hate speech has often focused on a single platform, commonly Twitter, but our goal was to study hate speech in a variety of settings and to ultimately develop a model to measure hate speech without reliance on platform-specific language patterns \citep{fortuna2018survey}.

To select comments for annotation, we used an enrichment sampling method to increase the relevance of the labeled comments to our theorized levels of hate speech, and in an effort to maximize the generalizability of our deep learning algorithm, we maintained a positive probability of selection for all sampled comments (i.e. no comments would be excluded based on their word usage).

Our sampling method relied on two dimensions for stratified sampling: 1) a relevance estimate of how likely the comment was to contain a target identity group, and 2) a hypothesis score for how hateful the comment was estimated to be. We used the identity relevance and hate speech hypothesis scores to create five stratification bins: 1) irrelevant (i.e. estimated to contain no references to identity groups), 2) relevant and low on predicted hate speech score (potential counterspeech or positive identity speech), 3) relevant and moderate on predicted hate speech score (neutral), 4) relevant and high on predicted hate speech score (low or moderate intensity hate speech), and 5) relevant and very high on predicted hate speech score (violent hate speech). We heavily oversampled bins 2, 4, and 5, and undersampled bins 1 and 3. As in a case-control study, this enriched sample could be re-weighted back to the original population of comments through inverse probability weighting \citep{horvitz1952generalization}.\footnote{See \citep[Appendix C]{delisle2019large} for a formal description of such re-weighting analysis.} We incorporated platform sample size targets such that our annotation dataset consisted of 40\% of the comments being sourced from Reddit, 40\% from Twitter, and 20\% from YouTube.

\subsection{Annotation}

Human reviewers were recruited from Amazon Mechanical Turk to complete our labeling instrument hosted on the Qualtrics survey platform between August 2, 2019 and September 10, 2019. Each annotator was given 26 comments---6 reference set comments and 20 "original comments"---to label. Participants were compensated \$7 and had a median completion time of 49 minutes, which resulted in a median pay rate of \$8.57 per hour. Study procedures were approved by the University of California, Berkeley Institutional Review Board.

\subsection{Psychometric scaling}

We applied faceted Rasch item response theory \citep{linacre1989facet}, a form of psychometric scaling designed for annotator evaluations, to combine the 10 ordinal components of hate speech into an interval-valued continuous hate speech score - our primary outcome.  With the Rasch faceted partial credit model the probability of the observed label for all hate speech components can be jointly modeled with maximum likelihood using the following multilevel equation \citep{linacre2002construction, eckes2015introduction}, which yields the interval hate speech score $\theta_n$: 
\begin{equation}
    \log \Big[ \frac{p_{nijk}}{p_{nijk-1}} \Big] = \theta_n - \delta_i - \alpha_j - \tau_{k}
\end{equation}
where:
\begin{itemize}
    \item $p_{nijk}$ is the probability of comment $n$ being rated as label $k$ by rater $j$ on component $i$,
    \item $p_{nijk-1}$ is the probability of comment $n$ being rated as label $k - 1$ by rater $j$ on component $i$,
    \item $\theta_n$ is the hate speech score of comment $n$ (i.e. a random effect),
    \item $\delta_i$ is the difficulty of component $i$,
    \item $\alpha_j$ is the perspective (``severity'') of rater $j$,
    \item $\tau_{k}$ is the difficulty of receiving label $k$ relative to label $k - 1$.
\end{itemize}

We used Facets software \cite{linacre2019facets} to conduct the many-facet Rasch scaling while leveraging IRT fit statistics and other quality indicators to remove raters estimated to have provided poor-quality labels; details are provided in the supplemental information.

\section{Modeling}

\begin{figure}[ht!]
\includegraphics[width=1\columnwidth]{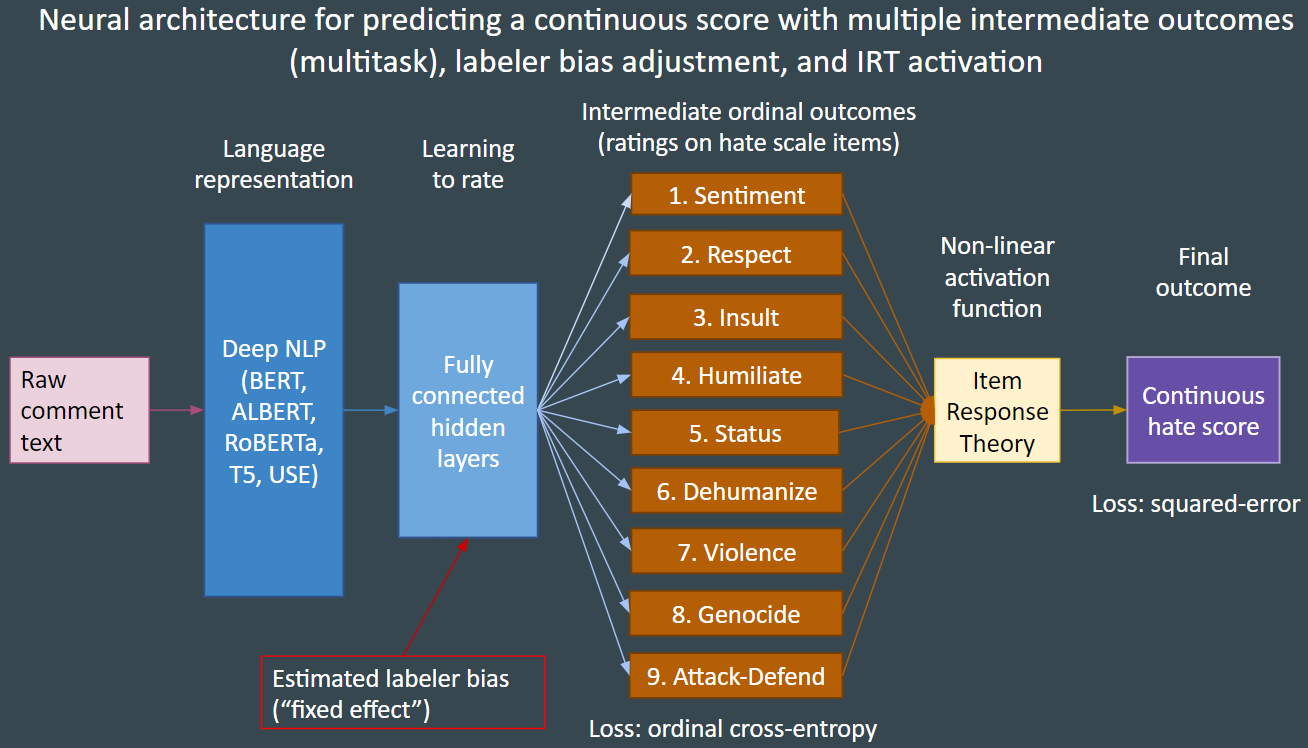}
\caption{Neural architecture for predicting a continuous score with multiple intermediate outcomes, labeler bias adjustment, and IRT activation.}
\label{fig-deep-irt}
\end{figure}

We then conducted supervised learning using the raw comment text as our input data and the latent hate speech score ($\theta_n$ above) as the outcome. We made four novel changes in our primary supervised approach: 1) rather than predict the hate speech score directly, we instead predicted the label for each component using a multitask architecture (i.e. multiple outputs within a single model) \cite{ruder2017overview}, a type of concept bottleneck model \cite{koh2020concept}, 2) each component was directly analyzed as an ordinal label using the consistent rank logits method of ordinal softmax activation and ordinal cross-entropy loss \cite{cao2019rank}, 3) we supplied the rater's estimated annotation instrument perspective (severity, $\alpha_j$ in equation 1) as an additional non-text input to allow the model to adjust its understanding of the likely item response, and 4) we tagged known slurs in the raw text of the comments, giving the deep learning models comparable information to what the human annotators were provided (and the potential to dynamically improve the slur tagging over time without retrained the model, as coded language evolves). The model's predictions on the 10 components of hate speech could then be batch-transformed into latent scores using the estimated IRT parameters from equation 1. See Figure \ref{fig-deep-irt} 
for a depiction of this architecture. We also tested the direct modeling of the continuous hate speech score as a simpler option, and of modeling the components as categorical rather than ordinal variables.
 
\section{Results}

\subsection{Psychometric scaling of ordinal labels}

\label{results-irt}

\begin{figure}[!ht]
\includegraphics[width=1\columnwidth]{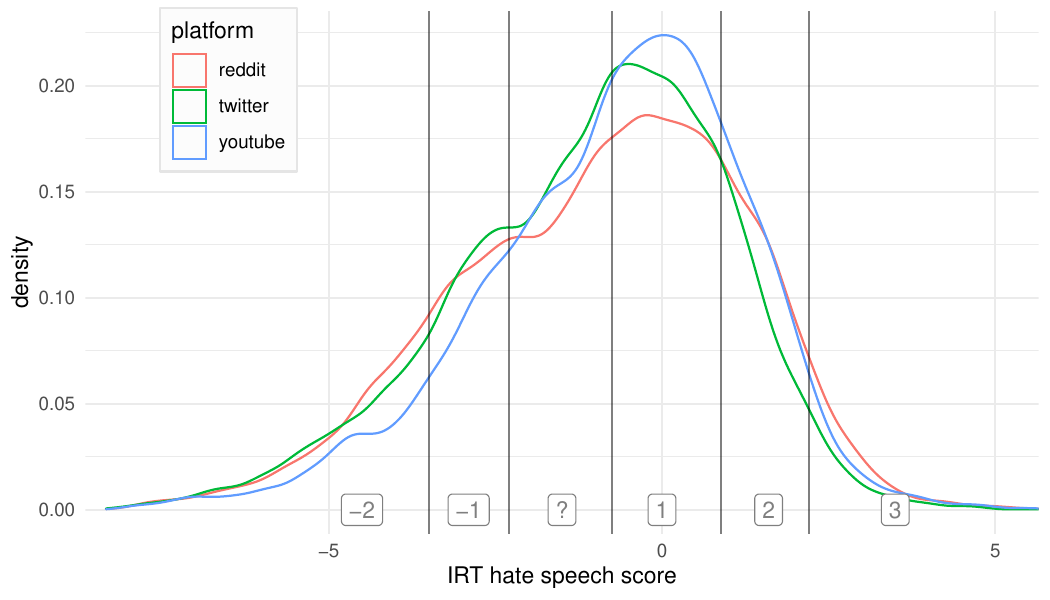}
\caption{Distribution of hate speech scores by social media platform, with vertical lines displaying proposed thresholds separating the revised theoretical construct levels.}
\label{irt-score-platform}
\end{figure}

Our faceted partial credit model achieved a reliability of 94\% for the hate speech scores, suggesting that 5-6 distinct strata can be identified across the continuous spectrum.\footnote{\url{https://www.winsteps.com/winman/reliability.htm}} 
Our estimated rater separation reliability was also 94\%, suggesting that our annotation design resulted in high accuracy at estimating the individual perspective (severity) for the annotators.  
Examining the hate speech scores across platforms, we found relatively similar distributions (Figure \ref{irt-score-platform}).

\begin{figure}[ht!]
\includegraphics[width=1\columnwidth]{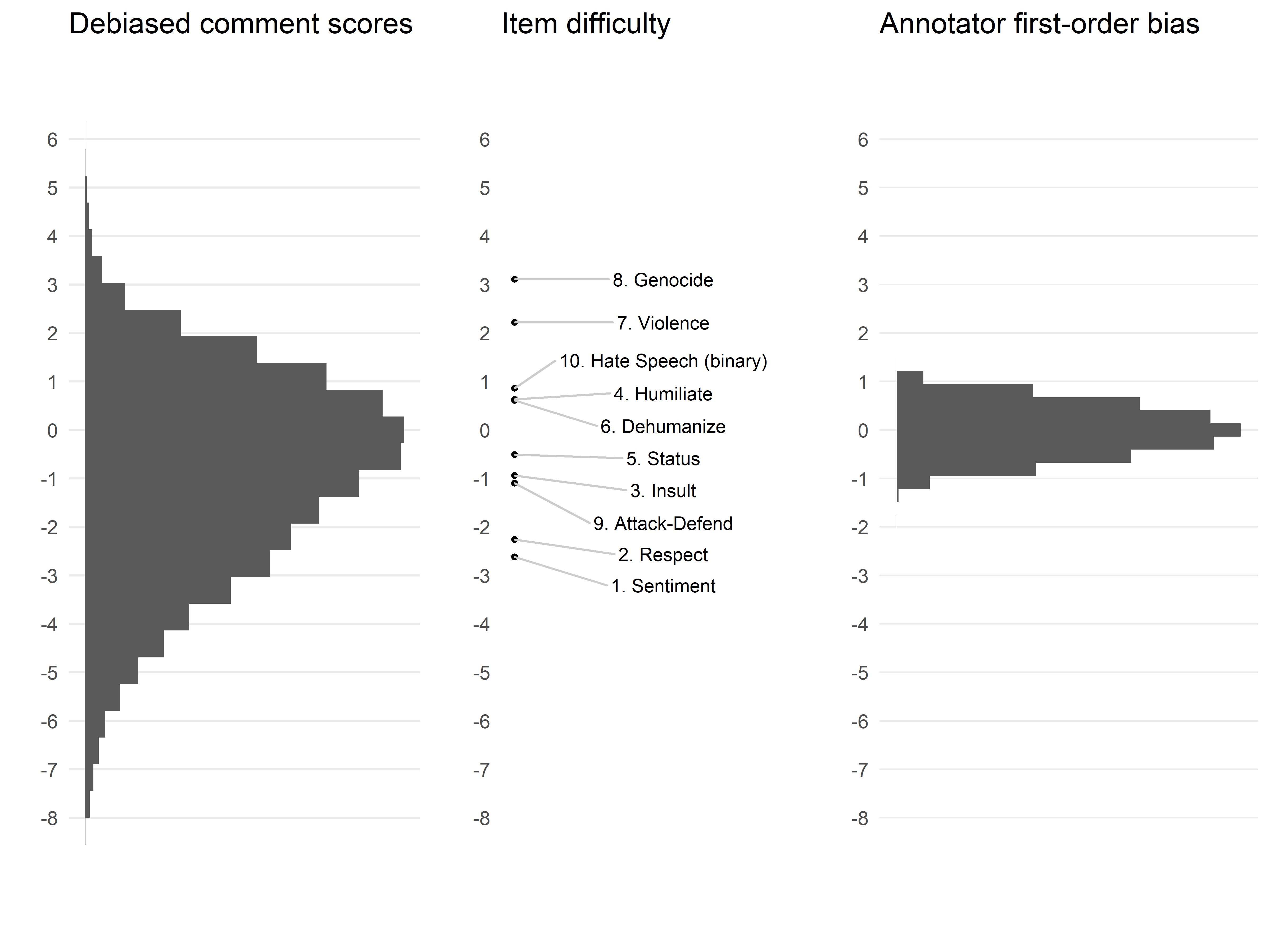}
\caption{Calibrated Wright map showing the comment hate scores ($\theta_n$),  component difficulty estimates ($\delta_i$), and annotator perspective estimates ($\alpha_j$).
}
\label{fit-wright-map}
\end{figure}

A calibrated Wright Map \citep{wilson2004constructing} showing our parameter estimates for the annotated data across comments ($\theta_n$), components / items ($\delta_i$), and annotators ($\alpha_j$) is shown in Figure \ref{fit-wright-map}. The difficult of a component shows where on the hate speech spectrum the component provides the most information. Additional details on the model fit statistics and psychometric characteristics are provided in the supplement.

\subsection{Supervised deep learning}

After IRT scaling, we used supervised deep learning to estimate a mapping from the raw text to the continuous hate speech score. Our current best models used a RoBERTa-Large contextual language base to process the raw text into a vector representation \citep{liu2019roberta}, followed by a 64-unit hidden layer with 10\% dropout, and included fine-tuning of the language representation subnetwork using HuggingFace software's Keras implementation \citep{wolf2019huggingface, chollet2015keras}. Results from our model testing are shown in Table \ref{results-performance}. Our proposed multitask networks that are transformed via IRT achieved the best performance, although we did not find a benefit from ordinal modeling of the items compared to simpler categorical modeling.

\begin{table}[!ht]
\centering
\caption{{\bf Supervised learning results.}. Corr = linear (Pearson) correlation. MAE = mean absolute error.}
\label{results-performance}
\begin{tabular}{|p{2.7cm}|
p{1.8cm}|p{0.8cm}|p{0.8cm}|}
\hline
\textbf{Algorithm} & 
\textbf{Outcome} & 
\textbf{MAE} &
\textbf{Corr.} \\ \hline
Jigsaw Identity Attack & Continuous & 1.914 & 0.423 \\ \hline
Jigsaw Toxicity & Continuous
& 1.904 & 0.655 \\ \hline
BERT-Large & Continuous  & 0.893 & 0.824  \\ \hline
RoBERTa-Large & Continuous & 0.861
& 0.839 \\ \hline
RoBERTa-Large & Ordinal + IRT & 0.858
& 0.841 \\ \hline
RoBERTa-Large & Categorical + IRT & \textbf{0.849}
& \textbf{0.843} \\ \hline
\end{tabular}%
\label{tab-deep-results}

\end{table}


\section{Conclusions}

We believe that this approach represents a novel approach to integrating deep learning with item response theory for measuring phenomena that can be decomposed into multiple items that are reviewed by human raters. It bears some similarity to concept bottleneck models \cite{koh2020concept}, though distinct theoretical motivation.

Faceted Rasch models include a noteworthy implication for inter-rater (kappa) reliability: it is not essential that different raters provide the same responses to an item when analyzing a certain comment. There is a growing body of related literature showing that a single "true" labeled response is often unrealistic \cite{aroyo2019crowdsourcing, palomaki2018case, geva2019we, pavlick2019inherent}. Instead, we want within-rater consistency in item interpretation so that their estimated severity acts as a strong summary measure of their individual style of rating. In fact, for raters with very different estimated severities, we would expect them to provide different ratings on an item when analyzing the same comment - that would be consistent with the measurement model (and common sense). Reliability of ratings and unbiasedness are two distinct phenomena; for example, a given comment may exhibit high reliability due to multiple raters agreeing on an biased assessment \cite{henning1996accounting}. This is a marked psychometric departure from prior studies of hate speech or other supervised natural language processing topics, which have commonly relied on inter-rater reliability as the primary quality metric for dataset labeling \cite{ross2016hatespeech}.




\section*{Acknowledgements}
We thank Nora Broege for her contributions to this project and Mark Wilson for his helpful comments.

\bibliographystyle{acl_natbib}
\bibliography{anthology,custom}

\appendix

\section{Appendix}
\label{sec:appendix}

\subsection{Networked design of annotation}

\begin{figure}[!htbp]
\includegraphics[width=1\columnwidth]{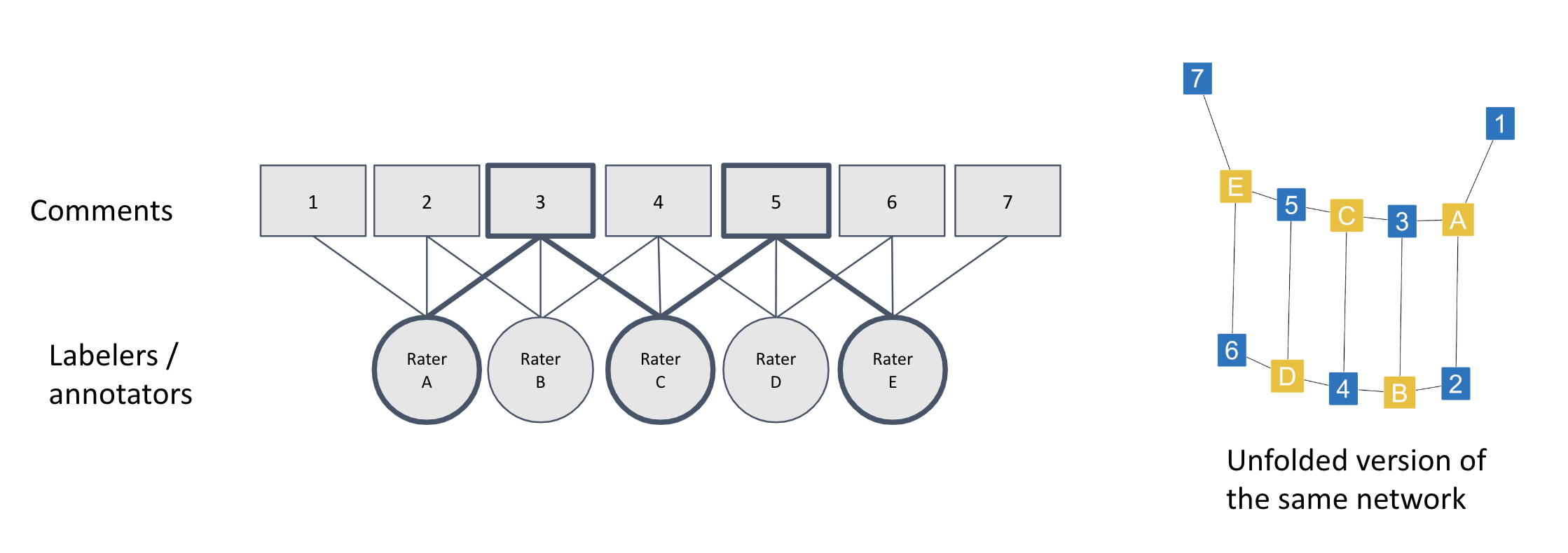}
\caption{Simplified annotation example showing network linkage across all reviewers.} 
\label{fig:network-example}
\end{figure}

Our comment annotation procedure incorporated a more complex network-based allocation of comments to annotators in order for the annotator perspective parameter to be statistically identifiable (i.e. estimable). Sampled comments were compiled into groups of 4 "original comments", stratified across our hypothesis score and relevance bins so that each group contained comments distributed across the hypothesized hate speech spectrum. Each comment group was randomly allocated to 4 comment batches to ensure 4 annotator reviews per comment, and each batch also included 6 reference set comments stratified across our 6 reference set levels. This design was chosen to generate a single network across all raters, which were linked through the comment groups plus the random selection from the reference set. In Figure \ref{fig:network-example} we show a simplified example of overlapping comment ratings that yield a single network across reviewers. This design further ensured that every reviewer would receive comments across our hate speech scale; we eliminated the risk that by random chance some raters would only review comments in a narrow range of the hypothesized scale.

\subsection{Correlation between component labels}

\begin{figure}[!ht]
\includegraphics[width=1\columnwidth]{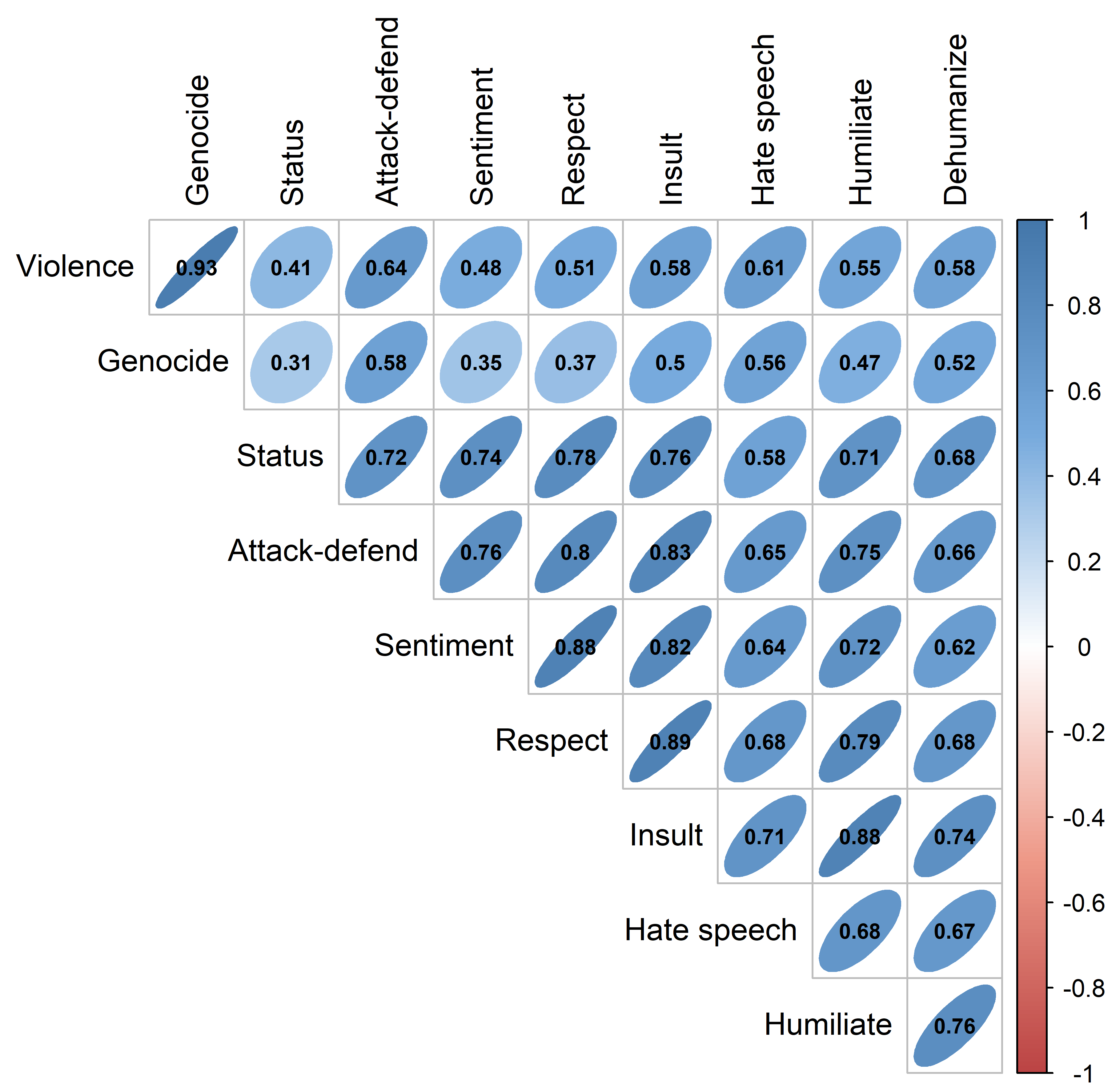}
\caption{Polychoric correlations between the items in our study.}
\label{item-correlations}
\end{figure}

A review of the correlation across component labels, shown in Figure \ref{item-correlations}, was suggestive that a multi-task deep learning architecture could benefit from sharing information across each item rating prediction. All items have a positive correlation with each other, varying from 0.31 to 0.93. The most highly correlated pair is genocide and violence (0.93), followed by insult and (dis)respect (0.89).

\subsection{Plausible value sampling}

\begin{figure}[!htbp]
\includegraphics[width=1\columnwidth]{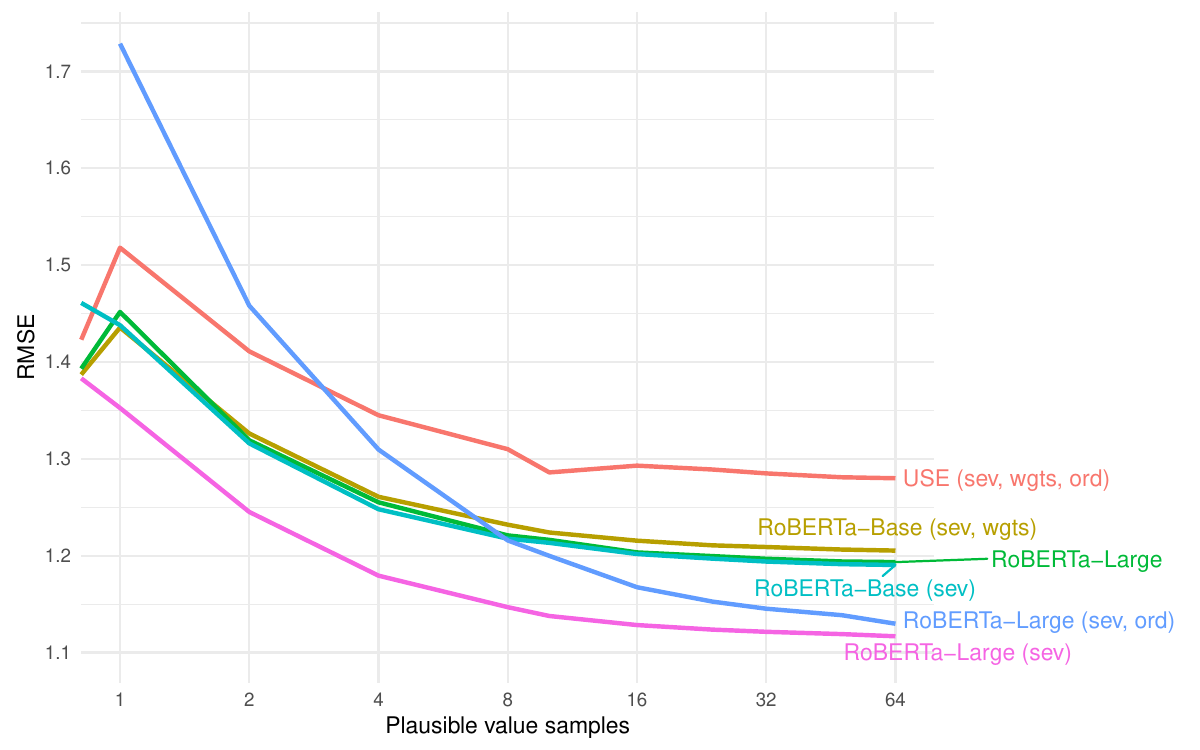}
\caption{Plausible value sampling improves model performance, up to a performance ceiling of approximately 64 samples, depending on model capacity. "Sev" = labeler severity included as an auxiliary input to the deep learning model. "Wgts" = observations weighted by inverse frequency, which deemphasizes the reference set comments. "Ord" = ordinal regression for item ratings, rather than categorical.}
\label{plausible-value-results}
\end{figure}

Multitask models benefited from plausible value sampling, which incorporated the probability distribution over item ratings, compared to predicting only the most probable item response for a given comment. Results are shown in Figure \ref{plausible-value-results}. Increasing the number of plausible valued improves the performance for all multitask models, but benefits begin to plateau at 16 or 32 plausible values, which strike a balance between improved accuracy and fast scaling time. There is suggestive evidence that ordinal models may benefit from larger plausible value sample sizes compared to categorical models, possibly due to improved calibration of the probability distributions.

\end{document}